\documentclass[10pt,twocolumn,letterpaper]{article}

\usepackage[pagenumbers]{cvpr} 

\newcommand{\myparagraph}[1]{\vspace{2pt}\noindent\textbf{#1}}

\usepackage{pdflscape}

\usepackage{xspace}
\usepackage{rotating}
\usepackage{amssymb,bm}

\usepackage{inconsolata}

\usepackage{subcaption}

\usepackage{multirow}

\usepackage[symbol]{footmisc}

\definecolor{cvprblue}{rgb}{0.21,0.49,0.74}
\usepackage[pagebackref,breaklinks,colorlinks,allcolors=cvprblue]{hyperref}

\def\paperID{83} 
\def\confName{CVPR}
\def\confYear{2025}

\title{One-Minute Video Generation with Test-Time Training}

\author{
\fontsize{10}{12}\selectfont
        Karan Dalal\thanks{Joint first authors}$^{*4} $
        \hspace{.1em} Daniel Koceja$^{*2}$\hspace{0.1em} Gashon Hussein$^{*2}$ \hspace{0.1em} Jiarui Xu$^{*1,3}$ 
        \hspace{0.1em} Yue Zhao\thanks{Joint second authors}$^{\dagger 5}$ \hspace{0.1em} Youjin Song$^{\dagger 2}$ \\[0.4em]
\fontsize{10}{12}\selectfont
        Shihao Han$^{1}$ \hspace{0.1em} Ka Chun Cheung$^{1}$ \hspace{0.1em} Jan Kautz$^{1}$ \hspace{0.1em} Carlos Guestrin$^{2}$ 
        \hspace{0.1em} Tatsunori Hashimoto$^{2}$ \hspace{0.1cm} Sanmi Koyejo$^{2}$ \\[0.4em]
\fontsize{10}{12}\selectfont
        Yejin Choi$^{1}$ \hspace{.1em} Yu Sun$^{1,2}$ \hspace{.1em} Xiaolong Wang$^{1,3}$ 
        \\[0.4em]
\fontsize{10}{12}\selectfont
        $^1$NVIDIA \quad
        $^2$Stanford University \quad
        $^3$UCSD \quad
        $^4$UC Berkeley \quad
        $^5$UT Austin
}

\begin{document}

\twocolumn[{
    \maketitle
    \centering
    \includegraphics[width=\textwidth]{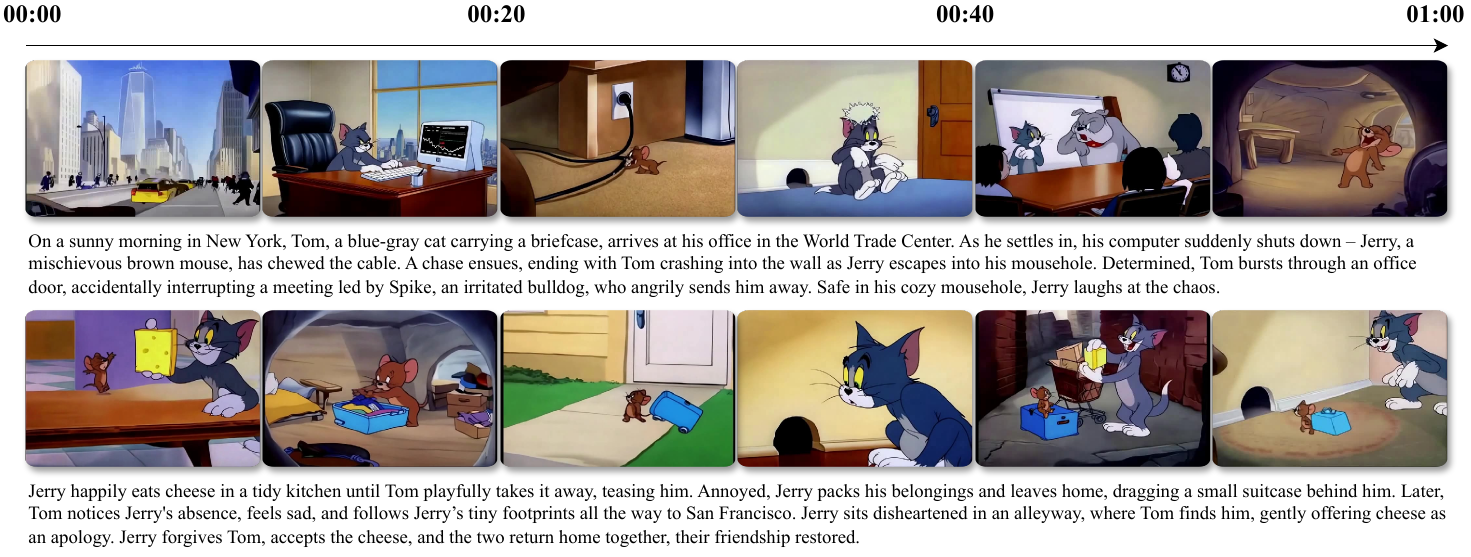}
    \captionof{figure}{TTT layers enable a pre-trained Diffusion Transformer to generate one-minute videos from text storyboards. 
    We use \textit{Tom and Jerry} cartoons as a proof of concept.
    The videos tell complex stories with coherent scenes composed of dynamic motion.
    Every video is produced directly by the model in a single shot, without editing, stitching, or post-processing. 
    Every story is newly created.
    }
    \label{fig:videos}
    \vspace{4ex}
}]

\footnotetext[1]{Joint first authors.~~$^\dagger$ Joint second authors.}
\footnotetext[0]{Accepted to The IEEE/CVF Conference on Computer Vision and Pattern Recognition (CVPR) 2025}

\renewcommand{\thefootnote}{\arabic{footnote}} 

\begin{abstract}

Transformers today still struggle to generate one-minute videos because self-attention layers are inefficient for long context.
Alternatives such as Mamba layers struggle with complex multi-scene stories because their hidden states are less expressive. 
We experiment with Test-Time Training (TTT) layers, whose hidden states themselves can be neural networks, therefore more expressive.
Adding TTT layers into a pre-trained Transformer enables it to generate one-minute videos from text storyboards.
For proof of concept, we curate a dataset based on \textit{Tom and Jerry} cartoons. 
Compared to baselines such as Mamba~2, Gated DeltaNet, and sliding-window attention layers, TTT layers generate much more coherent videos that tell complex stories, leading by 34 Elo points in a human evaluation of 100 videos per method.
Although promising, results still contain artifacts, likely due to the limited capability of the pre-trained 5B model.
The efficiency of our implementation can also be improved.
We have only experimented with one-minute videos due to resource constraints, but the approach can be extended to longer videos and more complex stories.

Sample videos, code and annotations are available at:
\url{https://test-time-training.github.io/video-dit}

\end{abstract}
\section{Introduction}
\label{sec:intro}

\begin{figure*}[t!]
    \centering
    \includegraphics[width=0.8\textwidth]{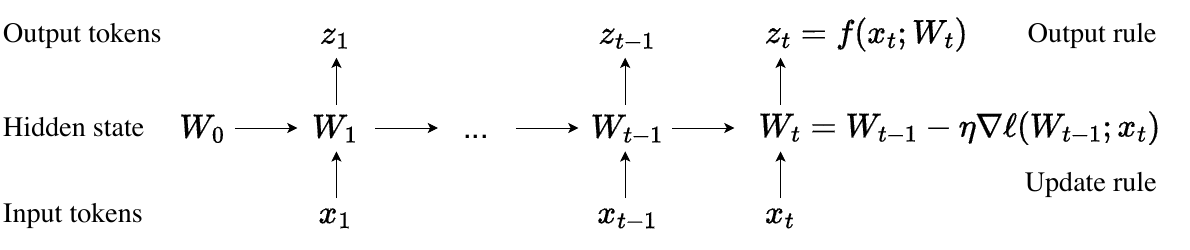}
    \caption{All RNN layers can be expressed as a hidden state that transitions according to an update rule.
    The key idea in \cite{sun2024ttt} is to make the hidden state itself a model $f$ with weights $W$, and the update rule a gradient step on the self-supervised loss $\ell$.
    Therefore, updating the hidden state on a test sequence is equivalent to training the model $f$ at test time. 
    This process, known as Test-Time Training (TTT), is programmed into TTT layers. 
    Figure and caption taken from \cite{sun2024ttt}.
    }
    \label{fig:ttt-layer}
\end{figure*}

Despite the remarkable progress in visual and physical realism, state-of-the-art video Transformers are still generating mostly short clips of single scenes without complex stories.
At the time of writing (March 2025), the maximum length of public APIs for video generation is 20 seconds for Sora (OpenAI), 16 seconds for MovieGen (Meta), 10 for Ray~2 (Luma), and 8 for Veo~2 (Google).
None of these APIs can autonomously generate complex multi-scene stories.

A fundamental challenge behind these technical limitations is long context, because the cost of self-attention layers in Transformers increases quadratically with context length.
This challenge is especially acute for video generation with dynamic motion, whose context cannot be easily compressed by a tokenizer.
Using a standard tokenizer, each of our one-minute videos requires over 300k tokens in context. 
With self-attention, generating a one-minute video would have taken $11\times$ longer than generating 20 videos of 3 seconds each, and training would have taken $12\times$ longer.

To address this challenge, recent work on video generation has investigated RNN layers as an efficient alternative to self-attention, because their cost increases linearly with context length~\cite{wang2024lingenhighresolutionminutelengthtexttovideo}.
Modern RNN layers, especially variants of linear attention~\cite{schmidhuberlinearattn, katharopoulos2020lineartransformers} such as Mamba~\cite{gu2024mamba, dao2024mamba2} and DeltaNet~\cite{schlag2021deltanet, yang2025gateddeltanetworksimproving}, have shown impressive results for natural language tasks.
However, we have yet to see long videos with complex stories or dynamic motion generated by RNNs.
Videos (\href{https://lineargen.github.io/}{link}) in \cite{wang2024lingenhighresolutionminutelengthtexttovideo} are high resolution and one-minute long, but contain only single scenes and slow motion, let alone complex stories.

We believe that these RNN layers generate less complex videos because their hidden states are less expressive.
RNN layers can only store past tokens into a hidden state of fixed size, which is only a matrix for linear attention variants such as Mamba and DeltaNet.
It is inherently challenging to compress hundreds of thousands of vectors into a matrix with only thousands in rank.
As a consequence, these RNN layers struggle to remember the deep relationships between distant tokens.

We experiment with an alternative class of RNN layers whose hidden states themselves can be neural networks. Specifically, we use two-layer MLPs with 2$\times$ more hidden cells and richer nonlinearities than the linear (matrix) hidden states in linear attention variants.
Since the neural network hidden states are updated by training even on test sequences, these new layers are called Test-Time Training (TTT) layers~\cite{sun2024ttt}.

We start from a pre-trained Diffusion Transformer (CogVideo-X 5B \cite{hong2023cogvideo}) that could only generate 3-second short clips at 16 fps (or 6 seconds at 8 fps).
Then, we add TTT layers initialized from scratch and fine-tune this model to generate one-minute videos from text storyboards. 
We limit the self-attention layers to 3-second segments so their cost stays manageable.
With only preliminary systems optimization, our training run takes the equivalent of 50 hours on 256 H100s.

We curate a text-to-video dataset based on $\approx$ 7 hours of \textit{Tom and Jerry} cartoons with human-annotated storyboards.
We intentionally limit our scope to this specific domain for fast research iteration.
As a proof-of-concept, our dataset emphasizes complex, multi-scene, and long-range stories with dynamic motion, where progress is still needed; it has less emphasis on visual and physical realism, where remarkable progress has already been made.
We believe that improvements in long-context capabilities for this specific domain will transfer to general-purpose video generation.

Compared to strong baselines such as Mamba 2~\cite{dao2024mamba2}, Gated DeltaNet~\cite{yang2025gateddeltanetworksimproving}, and sliding-window attention layers, TTT layers generate much more coherent videos that tell complex stories with dynamic motion, leading by 34 Elo points in a human evaluation of 100 videos per method.
For context, GPT-4o scores 29 Elo points over GPT-4 Turbo in LMSys Chatbot Arena~\cite{chiang2024chatbot}.

Sample videos, code and annotations are available at:
\url{https://test-time-training.github.io/video-dit}
\section{Test-Time Training Layers}
\label{sec:prelim}

Following standard practice~\cite{meta2024moviegen, yang2024cogvideox}, 
each video is pre-processed into a sequence of $T$ tokens, where $T$ is determined by its duration and resolution.
This section reviews Test-Time Training (TTT) layers for general sequence modeling, using some of the exposition in Section 2 of~\cite{sun2024ttt}.
We first discuss how to process general input sequences in a causal manner (chronological order).
Section~\ref{sec:method} discusses how to use RNN layers in a non-causal backbone by invoking them in opposite directions.

\subsection{TTT as Updating a Hidden State}
\label{subsec:hidden}
All RNN layers compress historical context in a hidden state of fixed size. This compression has two consequences.
On one hand, mapping an input token $x_t$ to output token $z_t$ is efficient, because both the update rule and output rule take constant time per token.
On the other hand, an RNN layer's ability to remember long context is limited by the amount of information its hidden state can store.
The goal of \cite{sun2024ttt} is to design RNN layers with expressive hidden states that can compress massive context.
As an inspiration, they observe that self-supervised learning can compress a massive training set into the weights of a machine learning model.

The key idea in \cite{sun2024ttt} is to use self-supervised learning to compress the historical context $x_1,\dots,x_t$ into a hidden state $W_t$, by making the context an unlabeled dataset and the hidden state the weights of a machine learning model $f$.
The update rule, illustrated in Figure~\ref{fig:ttt-layer}, is a step of gradient descent on some self-supervised loss $\ell$: 
\begin{equation}
\label{eq:update_naive}
W_t = W_{t-1} - \eta\,\nabla\ell(W_{t-1}; x_t),
\end{equation}
with learning rate $\eta$. Intuitively, the output token is just the prediction on $x_t$, made by $f$ with the updated weights $W_t$:
\begin{equation}
\label{eq:output_naive}
z_t = f(x_t; W_t).
\end{equation}

\noindent
One choice of $\ell$ is reconstructing $x_t$ itself. 
To make the learning problem nontrivial, one can first process $x_t$ into a corrupted input $\tilde{x}_t$ (see Subsection~\ref{subsec:task}), then optimize:
\begin{equation}    
\label{eq:recon}
\ell(W; x_t) = \| f(\tilde{x}_t; W) - x_t \|^2.
\end{equation}
Similar to denoising autoencoders~\citep{denoisingautoencoder}, $f$ needs to discover the correlations between dimensions of $x_t$ in order to reconstruct it from partial information $\tilde{x}_t$.

As with other RNN layers and self-attention, this algorithm that maps an input sequence $x_1,\dots,x_T$ to output sequence $z_1,\dots,z_T$ can be programmed into the forward pass of a sequence modeling layer.
Even at test time, the layer still trains a different sequence of weights $W_1, \dots, W_T$ for every input sequence. Therefore, it is called \emph{Test-Time Training (TTT) layer}.

Conceptually, calling backward on $\nabla\ell$ means taking gradients of gradients -- a well-explored technique in meta-learning.
TTT layers have the same interface as RNN layers and self-attention, therefore can be replaced in any larger network architecture. \cite{sun2024ttt} refers to training the larger network as the \emph{outer loop}, and training $W$ within each TTT layer as the \emph{inner loop}.

\subsection{Learning a Self-Supervised Task for TTT}
\label{subsec:task}
Arguably, the most important part of TTT is the self-supervised task specified by $\ell$. Instead of handcrafting a self-supervised task from human priors, \cite{sun2024ttt} takes a more end-to-end approach, learning it as part of the outer loop.
Starting from the naive reconstruction task in Equation~\ref{eq:recon}, they use a low-rank projection $\tilde{x}_t = \theta_Kx_t$, where $\theta_K$ is a matrix that is learnable in the outer loop.

Moreover, perhaps not all the information in $x_t$ is worth remembering, so the reconstruction label can also be a low-rank projection $\theta_Vx_t$ instead of $x_t$.
In summary, the self-supervised loss in \cite{sun2024ttt} is:
\begin{equation}
\label{eq:multi}
\ell(W; x_t) = \| f\left(\theta_K x_t; W\right) - \theta_V x_t \|^2.
\end{equation}
Lastly, since $\theta_Kx_t$ has fewer dimensions than $x_t$, \cite{sun2024ttt} can no longer use the output rule in Equation~\ref{eq:output_naive}.
So they make another projection $\theta_Qx_t$, and change the output rule to:
\begin{equation}
\label{eq:output}
z_t = f\left(\theta_Qx_t; W_t\right).
\end{equation}
Note that in the inner loop, only $W$ is optimized, therefore written as an argument of $\ell$; the $\theta$s are ``hyper-parameters" of this inner-loop loss function.
$\theta_K,\theta_V,\theta_Q$ are optimized in the outer loop, analogous to the Query, Key, and Value parameters of self-attention.

\subsection{TTT-MLP Instantiation}
Following \cite{sun2024ttt}, we instantiate the inner-loop model $f$ as a wrapper around $f_{\,\texttt{MLP}}$: a two-layer MLP similar to those in Transformers.
Specifically, the hidden dimension is $4 \times$ the input dimension, followed by a GELU activation~\cite{hendrycks2016gaussian}.
For better stability during TTT, $f$ always contains a Layer Norm and residual connection. That is,
$$f(x) = x + \texttt{LN}(f_{\,\texttt{MLP}}(x)).$$
A TTT layer with this $f$ is called TTT-MLP, which is the default instantiation throughout this paper.
In Section~\ref{sec:experiment} we also instantiate TTT-Linear (the $f$ above wrapping around a linear model) as a baseline.
\begin{figure*}[t!]
    \centering
    \includegraphics[width=\textwidth]{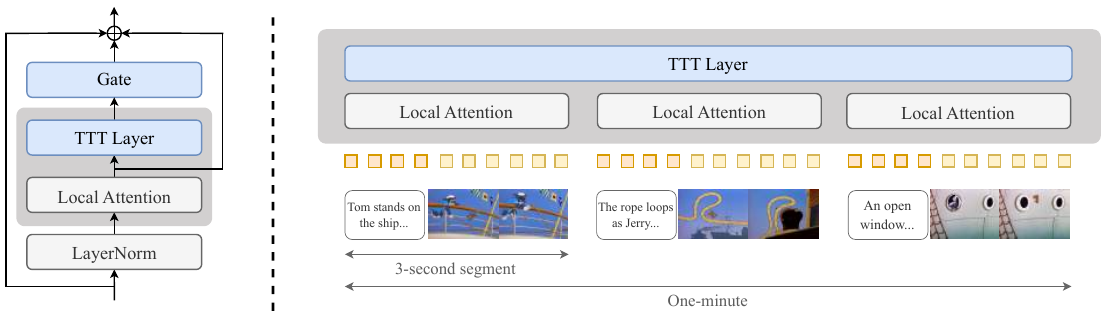}
    \caption{Overview of our approach. 
    \textbf{Left}: Our modified architecture adds a TTT layer with a learnable gate after each attention layer. See Subsection~\ref{subsec:arch}.
    \textbf{Right}: Our overall pipeline creates input sequences composed of 3-second segments. 
    This structure enables us to apply self-attention layers locally over segments and TTT layers globally over the entire sequence.
    See Subsection~\ref{subsec:pipeline}.}
    \label{fig:integration}
\end{figure*}

\section{Approach}
\label{sec:method}

At a high level, our approach simply adds TTT layers to a pre-trained Diffusion Transformer and fine-tunes it on long videos with text annotations. At a practical level, making this approach work involves many design choices.

\subsection{Architecture}
\label{subsec:arch}

\myparagraph{Pre-trained Diffusion Transformer.}
Our approach of adding TTT layers then fine-tuning can, in principle, work with any backbone architecture.
We choose Diffusion Transformers~\cite{peebles2023scalable} for our initial demonstration because it is the most popular architecture for video generation. 
Since the cost of pre-training a Diffusion Transformer on videos is prohibitive, we start from a pre-trained checkpoint called CogVideo-X 5B~\cite{hong2023cogvideo}.

\myparagraph{Gating.}
Given an input sequence $X = (x_1, \dots, x_T)$ where each token $x_t\in\mathbb{R}^d$, a TTT layer produces an output sequence $Z = (z_1, \dots, z_T) = \texttt{TTT}(X)$.
Each $z_t\in\mathbb{R}^d$ follows the recurrence described by Equations~\ref{eq:update_naive}, \ref{eq:multi} and \ref{eq:output} in Section~\ref{sec:prelim}.
Naively inserting TTT layers into a pre-trained network would dramatically worsen its predictions at the beginning of fine-tuning, when the TTT layers are randomly initialized.
To avoid this degradation, we gate \texttt{TTT} 
with a learned vector $\alpha\in\mathbb{R}^d$ following standard practice~\cite{alayrac2022flamingo}:
\begin{equation}
\texttt{gate}(\texttt{TTT}, X; \alpha) = \tanh(\alpha) \otimes \texttt{TTT}(X) + X,
\end{equation}
where $\tanh(\alpha)\in(-1, 1)^d$ is multiplied element-wise with each $z_t$ in $Z = \texttt{TTT}(X)$.
We initialize all values in $\alpha$ to $0.1$, so the values in $\tanh(\alpha)$ are close to 0 ($\approx 0.1$) at the beginning of fine-tuning.
This initialization of $\alpha$ allows $\texttt{TTT}$ to still contribute to $\texttt{gate}(\texttt{TTT}, X; \alpha)$ without significantly overwriting $X$.

\myparagraph{Bi-direction.}
Diffusion models, including CogVideo-X, are non-causal, meaning that an output token $z_t$ can condition on all of $x_1, \dots, x_T$ instead of only the past tokens $x_1,\dots, x_t$.
To use TTT layers in a non-causal manner, we apply a standard trick called bi-direction~\cite{mo2024scalingdiffusionmambabidirectional}. Given an operator $\texttt{rev}(X) = (x_T, \dots, x_1)$ that reverses $X = (x_1, \dots, x_T)$ in time, we define 
\begin{equation}
\texttt{TTT}'(X) = \texttt{rev}(\texttt{TTT}(\texttt{rev}(X))).
\end{equation}
Since \texttt{rev} is applied twice, $\texttt{TTT}'(X)$ is still in chronological order.
But the TTT layer inside it now scans through $X$ in reverse-chronological order.

\myparagraph{Modified architecture.}
Standard Transformers, including CogVideo-X, contain interleaving sequence modeling blocks and MLP blocks.
Specifically, a standard sequence modeling block takes an input sequence $X$ and produces
\begin{align}
X' &= \texttt{self\_attn}(\texttt{LN}(X))\label{eq:original}
\\
Y &= X' + X,
\end{align}
where \texttt{LN} is Layer Norm\footnote{Diffusion Transformers such as CogVideo-X use adaptive LN~\cite{peebles2023scalable}.} and $X' + X$ forms a residual connection.
We only modify the sequence modeling blocks, leaving everything else in the architecture unchanged. 
Each modified block, illustrated in the left panel of Figure~\ref{fig:integration}, continues from the $X'$ in Equation~\ref{eq:original} and produces
\begin{align}
Z &= \texttt{gate}(\texttt{TTT}, X'; \alpha),\label{eq:alpha}\\
Z' &= \texttt{gate}(\texttt{TTT}', Z; \beta),\label{eq:beta}\\
Y &= Z' + X.
\label{eq:modified}
\end{align}
Note that $\texttt{TTT}'$ only makes another call to \texttt{TTT}, so they share the same underlying parameters $\theta_K,\theta_V,\theta_Q$.
But for gating, Equation~\ref{eq:alpha} and~\ref{eq:beta} use different parameters $\alpha$ and $\beta$.

\subsection{Overall Pipeline}
\label{subsec:pipeline}
In this subsection, we discuss how to create the input sequence of tokens to our architecture and how each sequence is processed in segments.
Except for the first two text formats in the upcoming discussion, everything applies to both fine-tuning and inference.
Our pipeline is illustrated in the right panel of Figure \ref{fig:integration}.

\label{sec:method:pipeline}
\myparagraph{Scenes and segments.}
We structure our videos to contain multiple scenes,\footnote{
A scene is loosely defined as ``a part of a film in which the action happens in one place or is of one particular type.'' (Oxford Dictionary) 
} and each scene contains one or more 3-second \emph{segments}.
We use a 3-second segment as the atomic unit of text-to-video pairing for three reasons:
\vspace{0.2em}
\begin{itemize}[itemsep=0.2em]
\item The maximum length of generation for the original pre-trained CogVideo-X is 3 seconds.
\item The length of most scenes in the \textit{Tom and Jerry} episodes is at least 3 seconds.
\item Building a dataset with multiple stages (Subsection~\ref{subsec:dataset}) is most convenient given 3-second segments.
\end{itemize}

\myparagraph{Formats of text prompts.}
At inference time, a user can write the text prompt for a long video in any of the three formats listed below in the order of increasing detail.
See Figure~\ref{fig:prompts} in Appendix for examples of each format.
\vspace{0.2em}
\begin{itemize}[itemsep=0.2em]
    \item \textbf{Format 1}: A short summary of the plot in 5-8 sentences. Some of the examples are shown in Figure~\ref{fig:videos}.
    \item \textbf{Format 2}: A more detailed plot in roughly 20 sentences, with each sentence roughly corresponding to a 3-second segment. 
    Sentences can be labeled as belonging to certain scenes or groups of scenes, but these labels will be treated only as suggestions.
    \item \textbf{Format 3}: A storyboard. Each 3-second segment is described by a paragraph of 3-5 sentences, containing details such as background colors and camera movements. Groups of one or more paragraphs are strictly enforced as belonging to certain scenes with the keywords \mbox{\texttt{<scene start>}} and \texttt{<scene end>}.
\end{itemize}
The actual input to our text tokenizer is always in Format~3 during both fine-tuning and inference.
Conversion between the formats is performed by Claude 3.7 Sonnet in the order of $1\rightarrow2\rightarrow3$.\footnote{We observe that converting from Format 1 directly to Format 3 results in worse ability to follow the style of the human annotations in Format 3 in the fine-tuning dataset.}
For fine-tuning, our human annotations are already in Format 3, as discussed in Subsection~\ref{subsec:dataset}.

\myparagraph{From text to sequences.}
After the original CogVideo-X tokenizes the input text for each video, it concatenates the text tokens with noisy video tokens to form the input sequence to the Transformer.
To generate a long video, we apply the same procedure independently for each 3-second segment.
Specifically, given a storyboard in Format~3 with $n$ paragraphs, we first produce $n$ \emph{sequence segments}, each containing text tokens extracted from the corresponding paragraph followed by video tokens.
Then we concatenate all $n$ sequence segments together to form the input sequence, which now has interleaved text and video tokens.

\myparagraph{Local attention, global TTT.}
\mbox{CogVideo-X} uses self-attention layers to process the entire input sequence globally for each video of maximum length 3 seconds, but global attention becomes inefficient for long videos.
To avoid increasing the context length of self-attention layers, we make them local to each 3-second segment, attending to each of the $n$ sequence segments independently.\footnote{As an artifact of our pre-processing step, the sequence segments actually have an overlap of 1 latent frame (1350 tokens).}
The TTT layers process the entire input sequence globally because they are efficient in long context.

\begin{figure*}[t!]
    \centering
    \includegraphics[width=\textwidth]{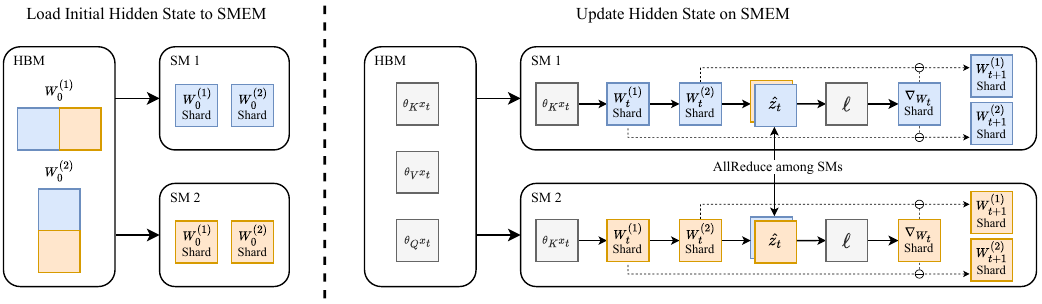}
    \caption{On-chip Tensor Parallel, discussed in Subsection~\ref{subsec:gpu}.
    \textbf{Left:} To reduce the memory required on each SM for TTT-MLP, we shard the hidden state $W^{(1)}$ and $W^{(2)}$ across SMs, transferring them between HBM and SMEM only during initial loading and final output.
    \textbf{Right:} We update the hidden state entirely on-chip and use the DSMEM feature on the NVIDIA Hopper GPU architecture to \texttt{AllReduce} intermediate activations among SMs.
    }
    \label{fig:sys}
\end{figure*}

\subsection{Fine-Tuning Recipe and Dataset}
\label{subsec:dataset}

\myparagraph{Multi-stage context extension.}
Following standard practice for LLMs~\cite{xiong2023effective}, we extend the context length of our modified architecture to one minute in five stages.
First, we fine-tune the entire pre-trained model on 3-second segments of \textit{Tom and Jerry} to adapt it to this domain.
New parameters (specifically those in TTT layers and gates) are assigned a higher learning rate during this stage.
Over the next four stages, we fine-tune on videos of 9, 18, 30, and eventually 63 seconds.
To avoid forgetting too much of the world knowledge from pre-training,
we only fine-tune the TTT layers, gates, and self-attention layers, using a lower learning rate during these four stages. 
See Appendix~\ref{sec:appendix:implementation} for the detailed recipe.

\myparagraph{Super-resolution on original videos.}
We start with 81 episodes of \textit{Tom and Jerry} released between 1940 and 1948.
Each episode is about 5 minutes, adding up to about 7 hours for all episodes.
The original videos vary in resolution, which is uniformly poor by modern standards.
We run a video super-resolution model~\cite{wang2021realesrgan} on the original videos, producing visually enhanced videos with shared resolution of $720\times480$ for our dataset.

\myparagraph{Multi-stage dataset.}
Following the structure discussed in Subsection~\ref{subsec:pipeline}, we first have human annotators break down each episode into scenes, then extract 3-second segments from each scene.
Next we have human annotators write a detailed paragraph for each 3-second segment.\footnote{
Each paragraph includes 1–2 sentences describing the background, 1–2 sentences describing the characters, and 2 sentences describing actions and camera movements. 
On average, each paragraph contains 98 words, which corresponds to 132 tokens.
}
Stage~1 fine-tunes directly on these segments.
To create data for the last four stages, we concatenate contiguous 3-second segments into videos of 9, 18, 30 and 63 seconds together with their text annotations.
Scene boundaries are marked by the same keywords in Subsection~\ref{subsec:pipeline}. As a result, annotations for all training videos are in Format~3.

\subsection{Parallelization for Non-Causal Sequences}
\label{subsec:parallel}
The update rule discussed in Section~\ref{sec:prelim} cannot be naively parallelized across tokens in a sequence, since computing $W_t$ requires $\nabla\ell(W_{t-1}; x_t)$, which in turn requires $W_{t-1}$.
To enable parallelization, we update $W$ on $b$ tokens at a time, which \cite{sun2024ttt} calls an inner-loop mini-batch.
Throughout this paper, we set $b=64$.

Concretely,  
for mini-batch $i = 1,\dots,T/b$ (assuming $T$ is an integer multiple of $b$), 
\begin{equation}
W_{ib} = W_{(i-1)b} - \frac{\eta}{b}\sum_{t=(i-1)b+1}^{ib}\nabla \ell\left(W_{(i-1)b}; x_t\right).
\label{eq:ttt_update_new}
\end{equation}
Because the sequence is non-causal, we then use $W_{ib}$ to produce the output tokens for all timesteps in mini-batch $i$:
\begin{equation}
    z_t = f(W_{ib}; x_t), \quad \quad \text{for}~~t = (i-1)b+1,\dots, ib.
    \label{eq:ttt_output_new}
\end{equation}
Note that $W_{(i-1)b+1},\dots,W_{ib-1}$ are no longer needed.

After this modification, $f$ can process an (inner-loop) mini-batch of tokens in parallel, similar to how a regular MLP processes an (outer-loop) mini-batch of training data. 
As a side benefit, we observe that averaging gradients across tokens reduces variance and stabilizes each update to $W$.

\subsection{On-Chip Tensor Parallel}
\label{subsec:gpu}
Implementing TTT-MLP efficiently for GPUs requires special designs to take advantage of their memory hierarchy.
A chip on a GPU is called a Streaming Multiprocessor (SM), analogous to a core on a CPU.
All SMs on a GPU share a relatively slow but large global memory called HBM, 
then each SM has a fast but small on-chip memory called SMEM.
Frequent data transfers between the SMEMs and HBM on a GPU can significantly hurt overall efficiency.

Efficient implementations of Mamba and self-attention layers (Flash Attention~\cite{dao2022flashattention}) use kernel fusion to minimize this kind of transfer. 
The high-level idea of these implementations is to load inputs and initial states into each SMEM, perform computations entirely on-chip, and write only the final outputs back to HBM. 
However, the hidden state for TTT-MLP, namely the weights $W^{(1)}$ and $W^{(2)}$ of the two-layer MLP $f$, is too large to be stored in the SMEM of a single SM (when combined with inputs and activations).

To reduce the memory required on each SM, we use Tensor Parallelism~\cite{shoeybi2019megatron} to shard $W^{(1)}$ and $W^{(2)}$ across SMs, as shown in Figure \ref{fig:sys}. 
Similar to how large MLP layers can be sharded and trained across the HBMs of multiple GPUs, 
we apply the same idea now across the SMEMs of multiple SMs, treating each SM as the analogy of a GPU. 
We use the DSMEM feature on the NVIDIA Hopper GPU architecture to implement \texttt{AllReduce} among SMs.
More details of our kernel are discussed in Appendix~\ref{sec:appendix:systems}.

Our implementation significantly improves efficiency, since hidden states and activations are now read from and written to HBMs only during initial loading and final output.
As a general principle, if a model architecture $f$ can be sharded with standard Tensor Parallelism across GPUs, then the same sharding strategy can be applied across SMs when $f$ is used as the hidden state.
\section{Evaluation}
\label{sec:experiment}

We perform human evaluation on a multi-axis benchmark for TTT-MLP and five baselines, all with linear complexity: local attention, TTT-Linear, Mamba 2, Gated DeltaNet, and sliding window attention layers.

\subsection{Baselines}
Except for local attention, all baselines are added to the same pre-trained CogVideo-X 5B using the approach in Subection~\ref{subsec:arch}; 
their modified architectures all have 7.2B parameters.
All baselines use the same fine-tuning recipe in Subsection~\ref{subsec:dataset}
and Appendix \ref{sec:appendix:implementation}.
Next we discuss the baselines in detail.
\vspace{0.2em}
\begin{itemize}[itemsep=0.2em]
\item\textbf{Local attention}: No modification to the original architecture, which performs self-attention on each 3-second segment independently.
\item\textbf{TTT-Linear}~\cite{sun2024ttt}: A TTT layer that instantiates $f(x) = x + \texttt{LN}(f_{\,\texttt{Linear}}(x))$, where $f_{\,\texttt{Linear}}$ is a linear model. 
\item\textbf{Mamba 2}~\cite{dao2024mamba2}: A modern RNN layer with a matrix hidden state, which is \(\approx4\times\) larger than the hidden state in TTT-Linear but \(\approx2\times\) smaller than that in TTT-MLP.
\item\textbf{Gated DeltaNet}~\cite{yang2025gateddeltanetworksimproving}: An extension of DeltaNet~\cite{yang2024parallelizing} and Mamba~2 with an improved update rule.
\item\textbf{Sliding-window attention}~\cite{beltagy2020longformerlongdocumenttransformer}: Self-attention with a fixed window of $8192$ tokens (about 1.5 seconds of video). 
\end{itemize}

\subsection{Evaluation Axes and Protocol}
\label{subsec:quan_eval}
From the six evaluation axes in MovieGen~\cite{meta2024moviegen}, we adopt the four relevant to our domain for human evaluation.\footnote{
Out of the six axes in MovieGen, we omit ``realness" which does not apply to cartoons.
We also omit ``motion completeness" which ``measures whether the output video contains enough motion", because all videos in our domain have highly dynamic motion.
We adapt ``frame consistency" to ``temporal consistency" to also include consistency across scenes.}

\begin{landscape}
\begin{figure}
    \centering
    \includegraphics[width=1.28\textwidth]{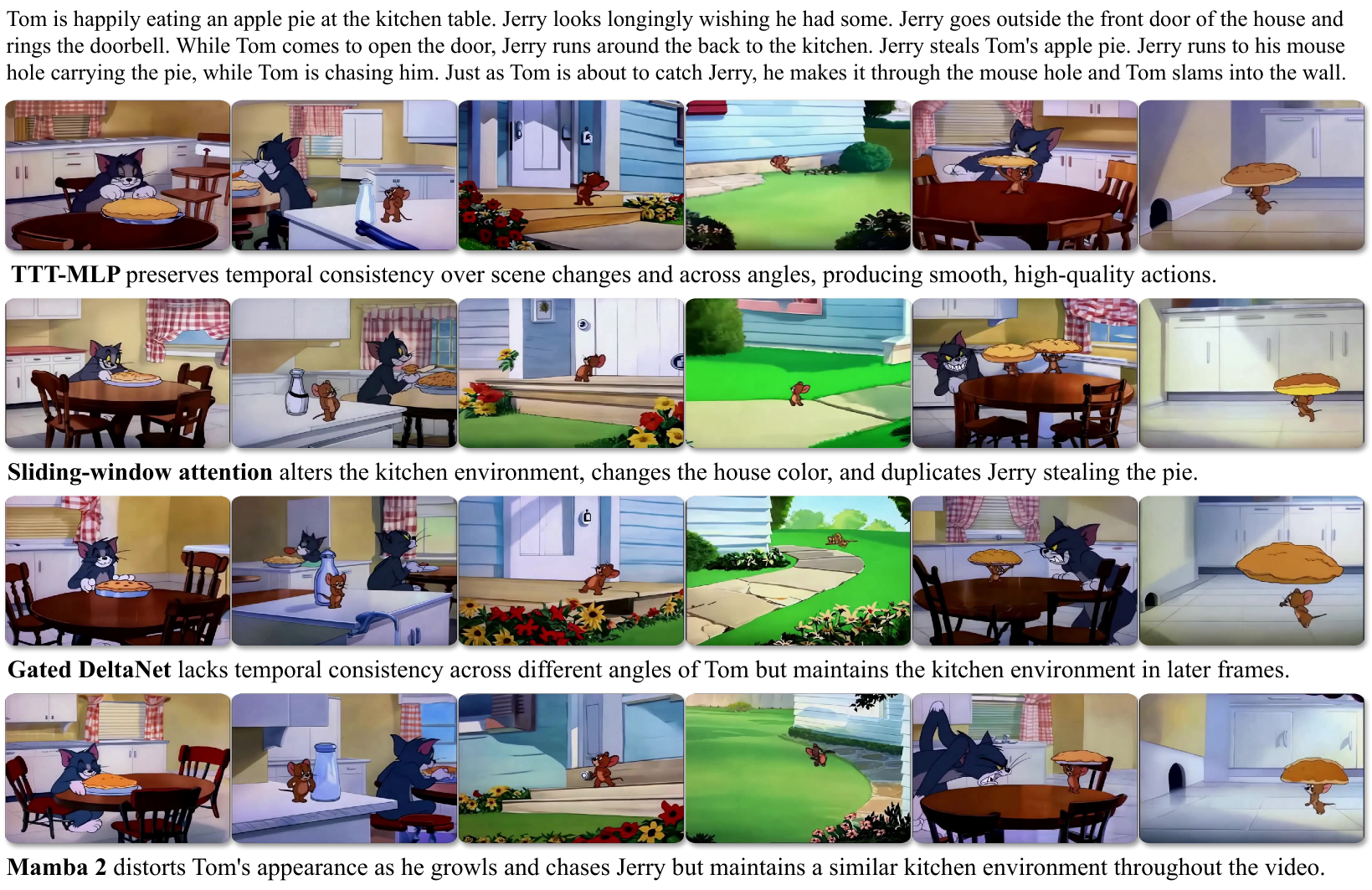}
    \caption{Video frames comparing TTT-MLP against Gated DeltaNet and sliding-window attention, the leading baselines in our human evaluation. 
    TTT-MLP demonstrates better scene consistency by preserving details across transitions and better motion naturalness by accurately depicting complex actions.}
    \label{fig:qualitative}
\end{figure}
\end{landscape}

\twocolumn[{
\centering
\setlength{\tabcolsep}{6pt}
\renewcommand{\arraystretch}{1.3}
\begin{tabular}{lcccc|c}
    \toprule
     & Text following & Motion naturalness & Aesthetics & Temporal consistency & Average \\
    \midrule
    {Mamba 2} & 985 & 976 & 963 & 988 & 978  \\
    {Gated DeltaNet} & 983 & 984 & 993 & 1004 & 991 \\
    {Sliding window} & \textbf{1016} & 1000 & 1006 & 975 & 999 \\
    {TTT-MLP} & 1014 & \textbf{1039} & \textbf{1037} & \textbf{1042} & \textbf{1033} \\
    \bottomrule
    \end{tabular}
    \captionof{table}{Human evaluation results for one-minute videos. TTT-MLP improves over the second best method by 34 Elo points on average. 
    Axes with the most improvements are scene consistency (+38) and motion smoothness (+39). For context, GPT-4 scores 46 Elo points over GPT-3.5 Turbo, and GPT-4o scores 29 over GPT-4 Turbo in Chatbot Arena~\cite{chiang2024chatbot}.}
    \label{tab:multiaxis_evaluation}

    \begin{minipage}[c]{0.66\textwidth}
    \vspace{2ex}
        \includegraphics[width=\linewidth]{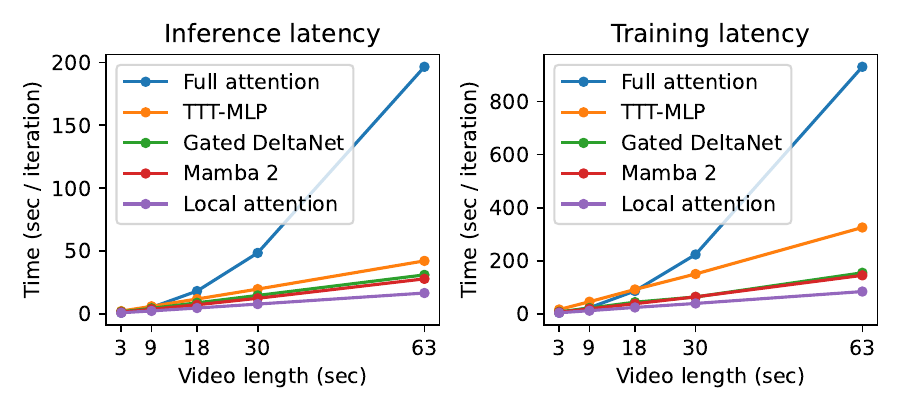}
    \end{minipage}\hfill
    \begin{minipage}[c]{0.33\textwidth}
    \vspace{2ex}
        \captionof{figure}{
        For 63-second videos, inference with full attention (over 300k tokens) would have taken $11\times$ longer than local attention, and training $12\times$ longer, as discussed in Section~\ref{sec:intro}.
        TTT-MLP takes $2.5\times$ and $3.8\times$ respectively -- significantly more efficient than full attention, but still less efficient than, for example, Gated DeltaNet, which takes $1.8\times$ longer than local attention in both inference and training.
        }
        \label{fig:your_label}
    \end{minipage}
    \vspace{1.0em}
}]

\begin{itemize}[itemsep=0.2em]
\item \textbf{Text following}: ``aligment with the provided prompt."
\item \textbf{Motion naturalness}: ``natural limb
movements, facial expressions, and adherence to physical laws.
Motion that appears unnatural or uncanny will be penalized."
\item \textbf{Aesthetics}: ``interesting and compelling content, lighting, color, and camera effects."
\item \textbf{Temporal consistency}: both inside and across scenes.
\end{itemize}
The quoted descriptions are from MovieGen~\cite{meta2024moviegen}.

Our evaluation is based on pairwise preferences in blind comparisons, because directly rating long videos or ranking many of them at once is challenging.
Specifically, an evaluator is given a random axis from the four above and a random pair of videos sharing the same plot, then asked to indicate the better video for that axis.
To collect the pool of videos, we first sample 100 plots using Claude 3.7 Sonnet (in Format $1\rightarrow2\rightarrow3$ as discussed in Subsection~\ref{subsec:pipeline}), then generate one video per method per plot.
The methods generating the videos are always unknown to the evaluators.

Our evaluators were recruited on \texttt{prolific.com} with the filters: living in the U.S., English as a first language, aged 18 to 35 years, with at least 100 previous submissions and an approval rate of at least 98\%.
The demographics of our evaluators, disclosed on the website, are as follows.
\vspace{0.2em}
\begin{itemize}[itemsep=0.2em]
\item \textbf{Gender}: 50.78\% male, 47.66\% female, 1.56\% other.
\item \textbf{Ethnicity}: 57.03\% White, 23.44\% Black, 10.94\% Mixed, 5.47\% Asian, and 3.12\% other. 
\end{itemize}
Based on this information, we believe that our evaluators constitute a representative sample of the U.S. population.

\begin{figure*}[!t]
    \centering
    \includegraphics[width=\textwidth]{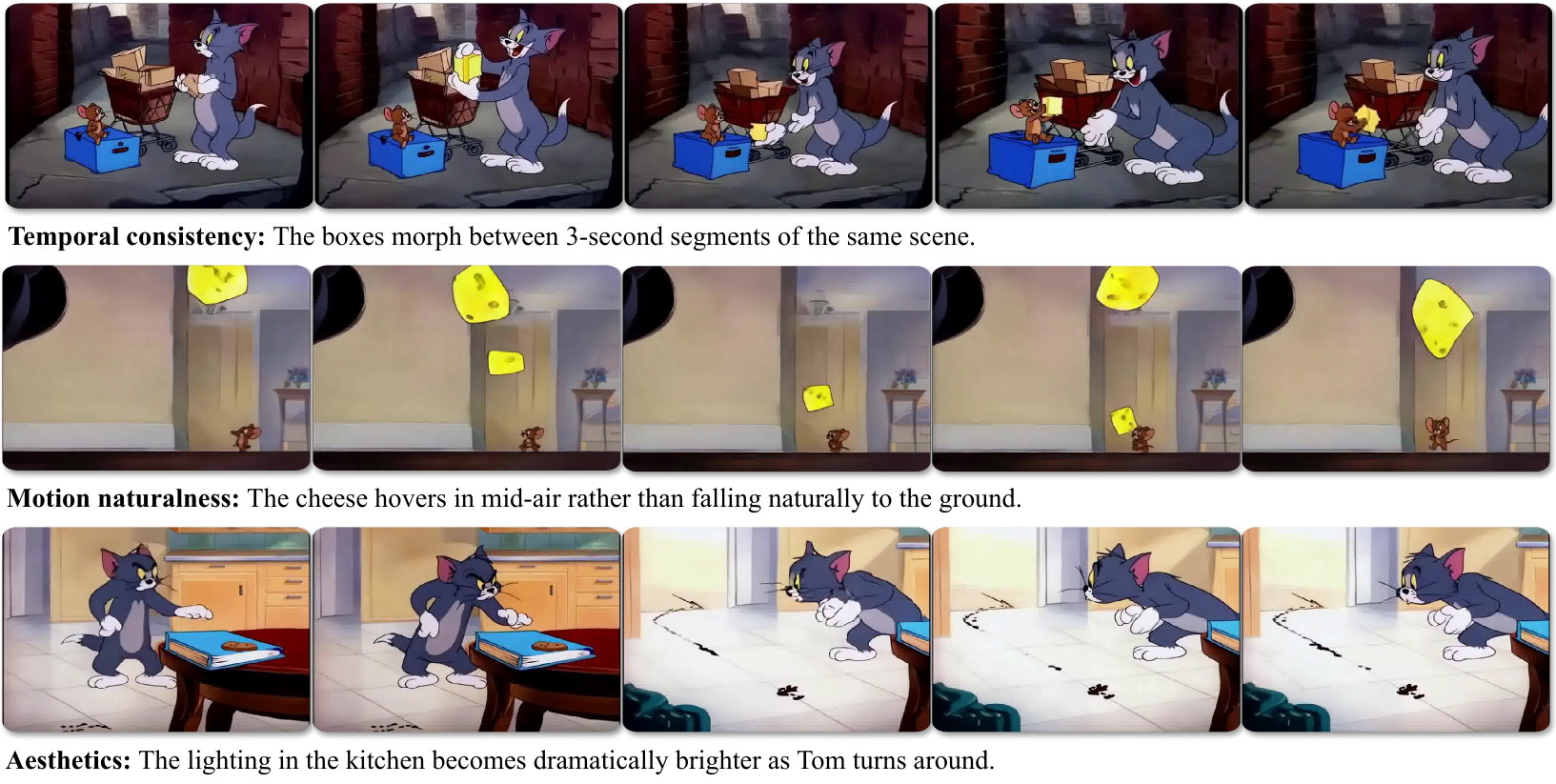}
    \vspace{-10pt}
    \caption{Artifacts in videos generated by TTT-MLP. 
    \textbf{Temporal consistency}: Objects sometimes morph at the boundaries of 3-second segments, potentially because the diffusion model samples from different modes across the segments.
    \textbf{Motion naturalness}: Objects sometimes float unnaturally because gravitational effects are not properly modeled.
    \textbf{Aesthetics}: Lighting changes do not consistently align with actions unless explicitly prompted. 
    Complex camera movements, such as parallax, are sometimes depicted inaccurately.
    }
    \label{fig:limitations}
\end{figure*}

\subsection{Results}
\label{subsec:results}

We aggregate the pairwise preferences using the Elo system in LMSys Chatbot Arena~\cite{chiang2024chatbot}.
The Elo scores are shown in Table~\ref{tab:multiaxis_evaluation}.

TTT-MLP improves over the second-best method by 34 Elo points on average. 
For context, GPT-4 scores 46 Elo points over GPT-3.5 Turbo (1163 vs. 1117), and GPT-4o scores 29 over GPT-4 Turbo (1285 vs. 1256) in LMSys Chatbot Arena~\cite{chiang2024chatbot}, so our improvement by 34 is practically meaningful.\footnote{
\url{https://lmarena.ai/}, accessed on March 20, 2025. The models considered are
GPT-4o-2024-05-13,
GPT-4-Turbo-2024-04-09,
GPT-4-0613, and
GPT-3.5-Turbo-0613.
}
Figure~\ref{fig:qualitative} compares frames of sample videos generated by TTT-MLP and the baselines.
The videos illustrated in Figure~\ref{fig:qualitative} can be accessed on the project website:
\url{https://test-time-training.github.io/video-dit}

\myparagraph{18-second elimination round.}
Note that local attention and TTT-Linear do not appear in Table~\ref{tab:multiaxis_evaluation}.
To avoid the much higher cost of evaluating longer videos on every method, we first conducted an elimination round using 18-second videos following the same procedure discussed in Subsection~\ref{subsec:quan_eval}.
This round eliminated local attention, which performed worst, and also TTT-Linear, which performed worse than TTT-MLP.
Results of the elimination round are shown in Table~\ref{tab:appendix:multiaxis_evaluation} in the Appendix.

\subsection{Limitations}
\label{subsec:limitations}
\myparagraph{Short context.}
For the 18-second elimination round discussed above, Gated DeltaNet performs the best on average, leading Mamba~2 by 27 Elo points and TTT-MLP by 28 (see Table~\ref{tab:appendix:multiaxis_evaluation} in the Appendix).
For 18-second videos, the context length is roughly 100k tokens.
This evaluation shows the scenario where RNN layers with linear (matrix) hidden states, such as Gated DeltaNet and Mamba~2, are still the most effective.
Moreover, evaluation results for both 18 and 63-second videos indicate that Gated DeltaNet improves meaningfully on Mamba~2.

\myparagraph{Wall-clock time.}
Even after applying our improvements in Subsection~\ref{subsec:parallel} and \ref{subsec:gpu}, the efficiency of TTT-MLP is still worse than Gated DeltaNet and Mamba~2.
This limitation is highlighted in Figure~\ref{fig:your_label}, where inference and training with TTT-MLP are $1.4\times$ and $2.1\times$ slower than with Gated DeltaNet, for example.
Section~\ref{sec:conclusion} discusses two potential improvements of our TTT-MLP kernel for better efficiency.
Note that training efficiency is not a significant concern in our application because the RNN layers are integrated after pre-training, which constitutes most of the overall training budget.
Training efficiency of the RNN layers is only relevant during fine-tuning, which is a small part of the budget to begin with.
In contrast, inference efficiency is much more meaningful.

\myparagraph{Video artifacts.}
The generated 63-second videos demonstrate clear potential as a proof of concept, but still contain notable artifacts, especially in motion naturalness and aesthetics. Figure~\ref{fig:limitations} illustrates examples of artifacts corresponding to three of our evaluation axes.
We observe that videos with these kinds of artifacts are not particular to TTT-MLP, but common among all methods.
The artifacts might have been a consequence of the limited capability of the pre-trained CogVideo-X 5B model.
For example,
videos (\href{https://github.com/THUDM/CogVideo}{link}) generated by the original CogVideo-X also seem to have limited motion naturalness and aesthetics.

\section{Related Work}
\label{sec:related}

\myparagraph{Modern RNN layers}, especially linear attention variants~\cite{schmidhuberlinearattn, katharopoulos2020lineartransformers}, such as Mamba~\cite{gu2024mamba, dao2024mamba2} and DeltaNet~\cite{schlag2021deltanet, yang2024parallelizing}, have demonstrated impressive performance in natural language tasks. 
Inspired by their success and ideas from Fast Weight Programmers~\cite{schmidhuber1992learning, kirsch2021meta, irie2021going, clark2022meta}, \cite{sun2024ttt} proposes scalable and practical ways to make the hidden states large and nonlinear, therefore more expressive.
Recent work~\cite{behrouz2024titans} develops even larger and more nonlinear hidden states, and updates them with more sophisticated optimization techniques.
The related work section in \cite{sun2024ttt} contains a detailed discussion of inspirations for TTT layers.
\cite{wang2025test} gives a good overview of recent developments in RNN layers.

\myparagraph{Long video modeling.}
Some early work~\cite{skorokhodov2022styleganv} generates long videos by training GAN~\cite{goodfellow2020gan,karras2020stylegan2} to predict the next frame based on the current frame and the motion vector.
Generation quality has improved significantly due to recent progress in auto-regression (AR) and diffusion-based approaches~\cite{gupta2024walt,meta2024moviegen,yang2024cogvideox,kong2025hunyuanvideo}.
TATS~\cite{ge2022tats} proposes the sliding window attention on the Transformer to generate videos longer than the training length.
Phenaki~\cite{villegas2023phenaki} works in a similar auto-regressive way, but each frame is generated by MaskGIT~\cite{chang2022maskgit}.
Pre-trained diffusion models can be extended to generate longer videos by using cascade~\cite{he2022lvdm,yin2023nuwa,wang2024lavie}, streaming~\cite{henschel2024streamingt2v}, and adding transitions~\cite{chen2023seine}.


\myparagraph{Story synthesis} methods such as \cite{li2019storygan,huang2016visual,pan2024synthesizing,maharana2022storydalle,rahman2023makeastory,liu2024storysalon} generate sequences of images or videos corresponding to individual sentences in a text story. 
For example, Craft~\cite{gupta2018flintstones} generates videos of complex scenes through retrieval, and StoryDiffusion~\cite{zhou2024storydiffusion} uses diffusion to improve the smoothness of transitions between frames. 
While related to text-to-video generation, story synthesis methods usually need additional components in their pipeline to maintain coherence across scenes, which are not processed end-to-end.

\section{Future Work}
\label{sec:conclusion}

We outline several promising directions for future work.

\myparagraph{Faster implementation.}
Our current TTT-MLP kernel is bottlenecked by register spills and suboptimal ordering of asynchronous instructions. Efficiency could probably be further improved by minimizing register pressure and developing a more compiler-aware implementation of asynchronous operations.

\myparagraph{Better integration.} Using bi-direction and learned gates is only one possible strategy for integrating TTT layers into a pre-trained model. Better strategies should further improve generation quality and accelerate fine-tuning. 
Other video generation backbones, such as autoregressive models, might require different integration strategies.

\myparagraph{Longer videos with larger hidden states.} 
Our approach can potentially be extended to generate much longer videos with linear complexity.
The key to achieving that goal, we believe, is to instantiate the hidden states as much larger neural networks than our two-layer MLP.
For example, $f$ itself can be a Transformer.

\vspace{4ex}
\myparagraph{Acknowledgements.} We thank Hyperbolic Labs for compute support, Yuntian Deng for help with running experiments, and Aaryan Singhal, Arjun Vikram, and Ben Spector for help with systems questions.
Yue Zhao would like to thank Philipp Krähenbühl for discussion and feedback.
Yu Sun would like to thank his PhD advisor Alyosha Efros for the insightful advice of looking at the pixels when working on machine learning.

\myparagraph{Note on authorship.} 
Gashon Hussein and Youjin Song joined the team after an initial version of this project was submitted to CVPR, and have made major contributions to the final version.
Because CVPR does not allow us to add authors after submission, their names could not appear on OpenReview and the conference webpage.
However, we all agree that the official author list should include their names, as presented in our released PDFs.
This project would not be possible without their work.

{
    \small
    \bibliographystyle{ieeenat_fullname}
    \bibliography{main}
}

\clearpage
\setcounter{page}{1}

\appendix

\twocolumn[{%
\centering
\setlength{\tabcolsep}{6pt}
\renewcommand{\arraystretch}{1.3}
\begin{tabular}{lccccc}
    \toprule
    Video len. & Ctx. len & Trainable parameters & Learning rate & Schedule & Steps \\
    \midrule
    3 sec & 18048 & TTT / Pre-trained Params & $1 \times 10^{-4}$ / $1\times 10^{-5}$ & Cosine / Constant & 5000 \\
    9 sec & 51456 & TTT + Local Attn (QKVO) & $1 \times 10^{-5}$ & Constant & 5000 \\
    18 sec & 99894 & TTT + Local Attn (QKVO) & $1 \times 10^{-5}$ & Constant & 1000 \\
    30 sec & 168320 & TTT + Local Attn (QKVO) & $1 \times 10^{-5}$ & Constant & 500 \\
    63 sec & 341550 & TTT + Local Attn (QKVO) & $1 \times 10^{-5}$ & Constant & 250 \\
    \bottomrule
\end{tabular}
\captionof{table}{Hyper-parameters for multi-stage fine-tuning. First, the entire pre-trained model is fine-tuned on 3-second segments of \textit{Tom and Jerry}, with higher learning rates assigned to the newly introduced TTT layers and gates. Then, only TTT layers, gates, and self-attention parameters are fine-tuned at reduced learning rates.}
\label{tab:appendix:hparams}
\vspace{1em}

\begin{tabular}{lcccc|c}
    \toprule
     & Text following & Motion naturalness & Aesthetics & Temporal consistency & Average \\
    \midrule
    {Local Attention} & 965 & 972 & 969 & 944 & 962 \\
    {TTT-Linear} & 1003 & 995 & 1007 & 1001 & 1001 \\
    {Mamba 2} & \textbf{1023} & 987 & 1008 & 1004 & 1005 \\
    {Gated DeltaNet} & 1020 & \textbf{1039} & \textbf{1044} & {1026} & \textbf{1032} \\
    {SWA} & 995 & 1004 & 993 & 980 & 993 \\
    {TTT-MLP} & 994 & 1002 & 1002 & 1019 & 1004 \\
    \bottomrule
\end{tabular}
\captionof{table}{Human evaluation results for 18-second videos, discussed in Subsection~\ref{subsec:results} and \ref{subsec:limitations}.}
\label{tab:appendix:multiaxis_evaluation}
\vspace{1em}
}]

\section{Experiment Details}
\label{sec:appendix:implementation}

\myparagraph{Diffusion schedule.} 
Following CogVideoX~\cite{yang2024cogvideox}, we fine-tune our model using v-prediction~\cite{salimans2022progressive}, which includes a diffusion noise schedule with 1000 steps and Zero-SNR~\cite{lin2024zerosnr} enforced at the final step.

\myparagraph{Training configurations.} We use the following hyper-parameters for all stages of training:

\vspace{0.2em}
\begin{itemize}[itemsep=0.2em]
    \item \textbf{Optimizer:} AdamW with $(\beta_1, \beta_2) = (0.9, 0.95)$
    \item \textbf{Learning Rate:} Linear warmup over 2\% of training steps
    \item \textbf{Batch Size:} 64
    \item \textbf{Gradient Clipping:} 0.1
    \item \textbf{Weight Decay:} $10^{-4}$ applied to all params except biases and normalization layers
    \item \textbf{VAE Scale Factor}: 1.0
    \item \textbf{Dropout:} Zero-out text prompt with probability 0.1
    \item \textbf{Precision:} Mixed Precision with PyTorch FSDP2
\end{itemize}

\myparagraph{TTT configurations.} A key hyperparameter for TTT layers is the inner-loop learning rate $\eta$, which we set $\eta = 1.0$ for TTT-Linear and $\eta = 0.1$ for TTT-MLP.

\myparagraph{Sampling schedule.} We follow the DDIM sampler~\cite{song2021ddim} with 50 steps, applying dynamic classifier-free guidance (CFG)~\cite{ho2022cfg} that increases CFG magnitude from 1 to 4 and utilizing negative prompts to further enhance video quality.

\section{On-Chip Tensor Parallel Details}
\label{sec:appendix:systems}

We use ThunderKittens~\cite{spector2025thunderkittens} to implement the TTT-MLP kernel, described in Subsection \ref{subsec:gpu}.

\vspace{0.75em} \myparagraph{Hidden state sharding.} We follow the standard strategy for Tensor Parallel, sharding the first layer column-wise and the second layer row-wise. As the GeLU non-linearity is elementwise, the forward pass of the TTT-layer requires a single reduction for computing the inner loss used to update the hidden state.

\myparagraph{Further latency optimizations.} We incorporate several techniques from FlashAttention-3 \cite{shah2024flashattention3fastaccurateattention} to further reduce I/O latency on NVIDIA Hopper GPUs. In particular, we implement a multi-stage pipelining scheme that asynchronously prefetches future mini-batches from HBM, overlapping data transfers with computation on the current mini-batch. This approach, known as \textit{producer-consumer asynchrony}, involves dedicating specialized warpgroups to either data loading (producer) or computation (consumer).

\myparagraph{Gradient checkpointing.}
We integrate gradient checkpointing along the sequence dimension~\cite{sun2024ttt} directly into our fused kernel. To reduce I/O-induced stalls and CUDA thread workloads, we use the Tensor Memory Accelerator (TMA) to perform asynchronous memory stores.

\begin{figure*}[!t]
    \centering
    \includegraphics[width=0.82\textwidth]{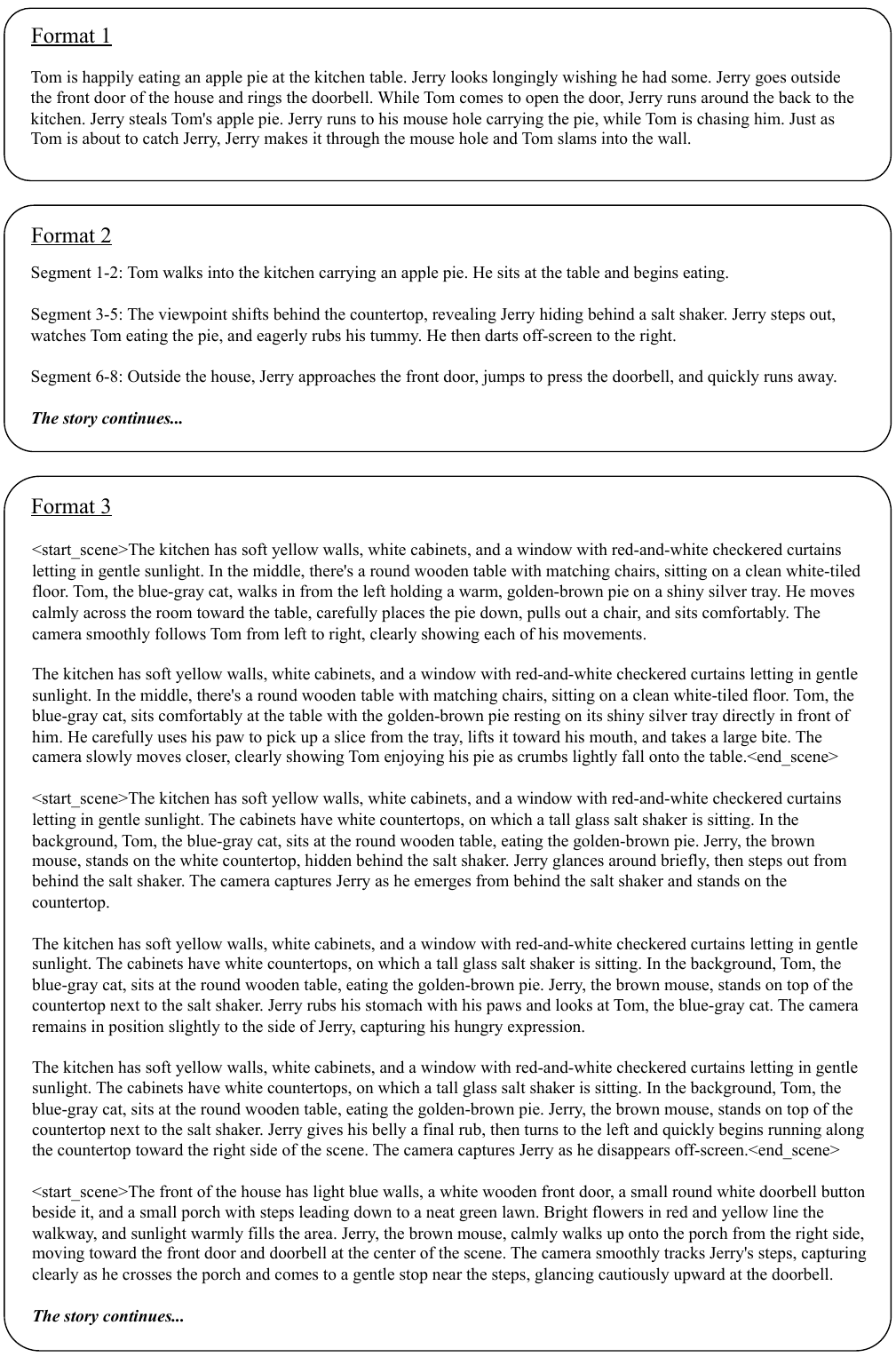}
    \caption{Illustration of the three prompt formats discussed in Subsection~\ref{subsec:pipeline}: (1) a short summary of the plot, (2) sentence-level descriptions of the segments, and (3) a detailed storyboard.}
    \label{fig:prompts}
\end{figure*}
\end{document}